\pdfoutput=1

\documentclass[11pt]{article}

\usepackage[]{acl}
\usepackage{graphicx}
\usepackage{caption}
\usepackage{subcaption}
\usepackage{tabularx}
\usepackage{amsmath}
\usepackage{amssymb}
\usepackage{booktabs}
\usepackage{pbox}
\usepackage{multirow}
\usepackage{xspace}

\usepackage{times}
\usepackage{latexsym}

\usepackage[T1]{fontenc}

\usepackage[utf8]{inputenc}

\usepackage{microtype}

\newcommand{\method}{FACTOR\xspace}
\newcommand{\wikimethod}{Wiki-FACTOR\xspace}
\newcommand{\newsmethod}{News-FACTOR\xspace}
\newcommand{\qamethod}{Expert-FACTOR\xspace}

\DeclareMathOperator*{\argmax}{argmax}

\title{Generating Benchmarks for Factuality Evaluation of Language Models}

\author{Dor Muhlgay\thanks{~~Corresponding author: \texttt{dorm@ai21.com}}~~~~Ori Ram~~~~Inbal Magar~~~~Yoav Levine~~~~Nir Ratner\\
\textbf{Yonatan Belinkov~~~Omri Abend~~~Kevin Leyton-Brown~~~Amnon Shashua~~~Yoav Shoham}\\\\
AI21 Labs}

\begin{document}
\maketitle

\maketitle
\begin{abstract} 

Before deploying a language model (LM) within a given domain, it is important to measure its tendency to generate factually incorrect information in that domain.
Existing methods for factuality evaluation of LLM generation focus on facts sampled from the LM itself, and thus do not control the set of evaluated facts and might under-represent domain specific or rare facts. 
We propose FACTOR: Factual Assessment via Corpus TransfORmation,
a scalable approach for evaluating LM factuality. 
\method automatically transforms a factual corpus of interest into a benchmark evaluating an LM's propensity to generate true facts from the corpus vs.\ similar but incorrect statements. We use our 
framework to create three benchmarks: \textit{Wiki-FACTOR}, \textit{News-FACTOR} and \textit{Expert-FACTOR}. We show that: 
(i) our benchmark scores increase with model size and improve when the LM is augmented with retrieval;    
(ii) benchmark score and perplexity do not always agree on model ranking; (iii) when perplexity and benchmark score disagree, the latter better reflects factuality in open-ended generation, as measured by human annotators. We make our data and code publicly available\footnote{\url{https://github.com/AI21Labs/factor}}. 

\end{abstract}
\section{Introduction}\label{sec:1}

\begin{figure}[t]
\centering
\includegraphics[width=1\columnwidth]{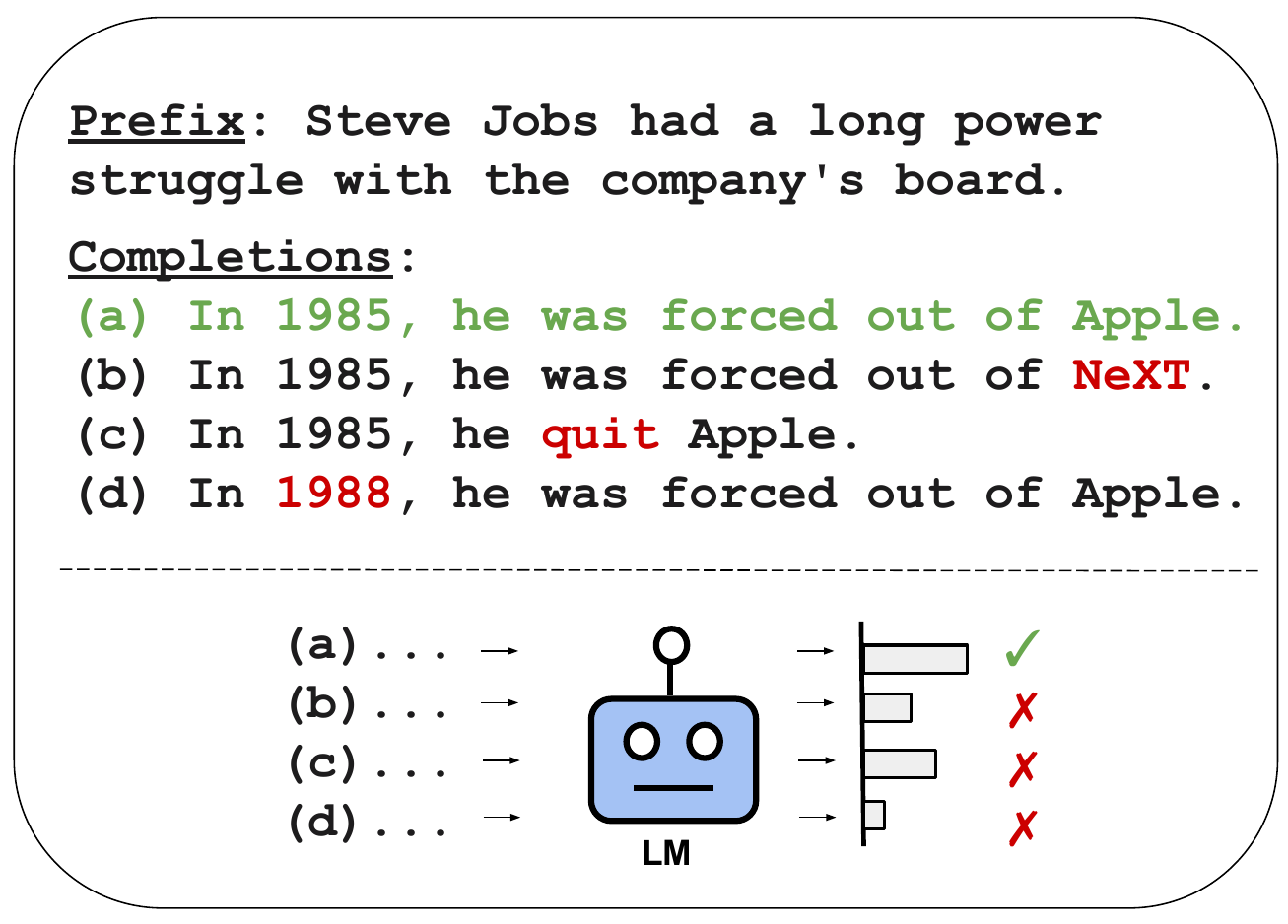}
\vspace{-12pt}
\caption{Each example in our evaluation task (dubbed \method) consists of a \textit{prefix} and four \textit{completions}, of which only one is factually correct (completion (a) in this example). The non-factual completions (b), (c) and (d), marked in red, are generated according to different factual error types, detailed in Table~\ref{tab:negxamples}. The evaluated model assigns likelihood scores to each completion separately. It is considered ``correct'' if it assigns the highest likelihood to the factually correct completion over all non-factual alternatives.
}

\label{fig:intro_fig}
\end{figure}

Despite rapid improvements in their capabilities, large Language Models (LMs) still tend to generate factually inaccurate or erroneous text~\cite{lin-etal-2022-truthfulqa,maynez-etal-2020-faithfulness,huang2020challenges}. 
Such phenomena can pose a significant hurdle to deploying LMs in important or sensitive settings, motivating the development of methods for evaluating LM factuality in open-ended generation. 

Methods for directly evaluating an LM's propensity towards factual generation were recently proposed by \citet{lee2022factuality} and \citet{min2023factscore}. These methods suggest sampling generations from a model, applying an automatic pipeline for fact verification, and then assigning a score corresponding to the percentage of factually correct generated statements. In task-specific domains, such as long-form question answering, evaluation is usually done by assessing the relevance of a sampled generation against a reference text \cite{lin-2004-rouge, fabbri-etal-2022-qafacteval}. However, the sampling approach may introduce bias: by scoring the accuracy of facts that an LM tends to generate in an open-ended setting, high-likelihood facts are over-represented, while the ``long-tail'' of rare facts is under-represented. 

Currently, there are no metrics suited to measuring LM factuality with respect to a \textit{controlled set of facts} in a \textit{generation setting}. A common proxy is measuring LM perplexity; this was widely adopted to evaluate retrieval-augmented LMs \cite{knn-lm,retro,ram2023context,shi2023replug}. However, perplexity is affected by many linguistic phenomena, and so cannot be directly linked to factuality.

This paper introduces a novel framework for testing a
model’s tendency to generate factual information
from a given factual corpus: Factual Assessment via Corpus TransfORmation (\textit{\method}). 
 The key idea is automatically perturbing factual statements taken from the corpus to create a constant number of similar but false variations for each true statement
 (Figure~\ref{fig:intro_fig}).
 We employed InstructGPT \cite{ouyang2022training} to generate the false variations for each true statement.
 The LM's \method accuracy on our benchmark is defined as the percentage of examples for which it assigns higher likelihood to the factual completion than to any of the false variations.  

We applied \method to the Wikipedia and News domains, as well as to a diverse collection of domain specific question-answer pairs (\textit{e.g.}, medicine, technology, law); constructing new benchmarks dubbed \textit{\wikimethod},  \textit{\newsmethod} and \textit{\qamethod}.
We used these datasets to evaluate a large suite of LMs from the OPT \cite{zhang2022opt}, GPT-2 \cite{gpt2}, and GPT-Neo \cite{gpt-neo} families, ranging from 110M to 66B parameters. 
We show in \S\ref{sec:results_size} that, as expected, \method scores increase with model size. However, even the largest models we evaluated achieved scores of only $58$\% for \wikimethod, $68$\% for \newsmethod, and $55$\% for \qamethod, indicating that these benchmarks are challenging even for large LMs. In \S\ref{sec:results_retrieval} we show that consistent \method score improvements can be achieved by augmenting the LMs with the simple retrieval component used by~\citet{ram2023context}. This directly demonstrates that retrieval augmentation improves factuality in the LM setting; 
\method is thus posed as a prominent approach for measuring retrieval-augmented LMs.

\begin{figure}[t]
\centering
\hspace*{-10pt}
\includegraphics[width=1.05\columnwidth]{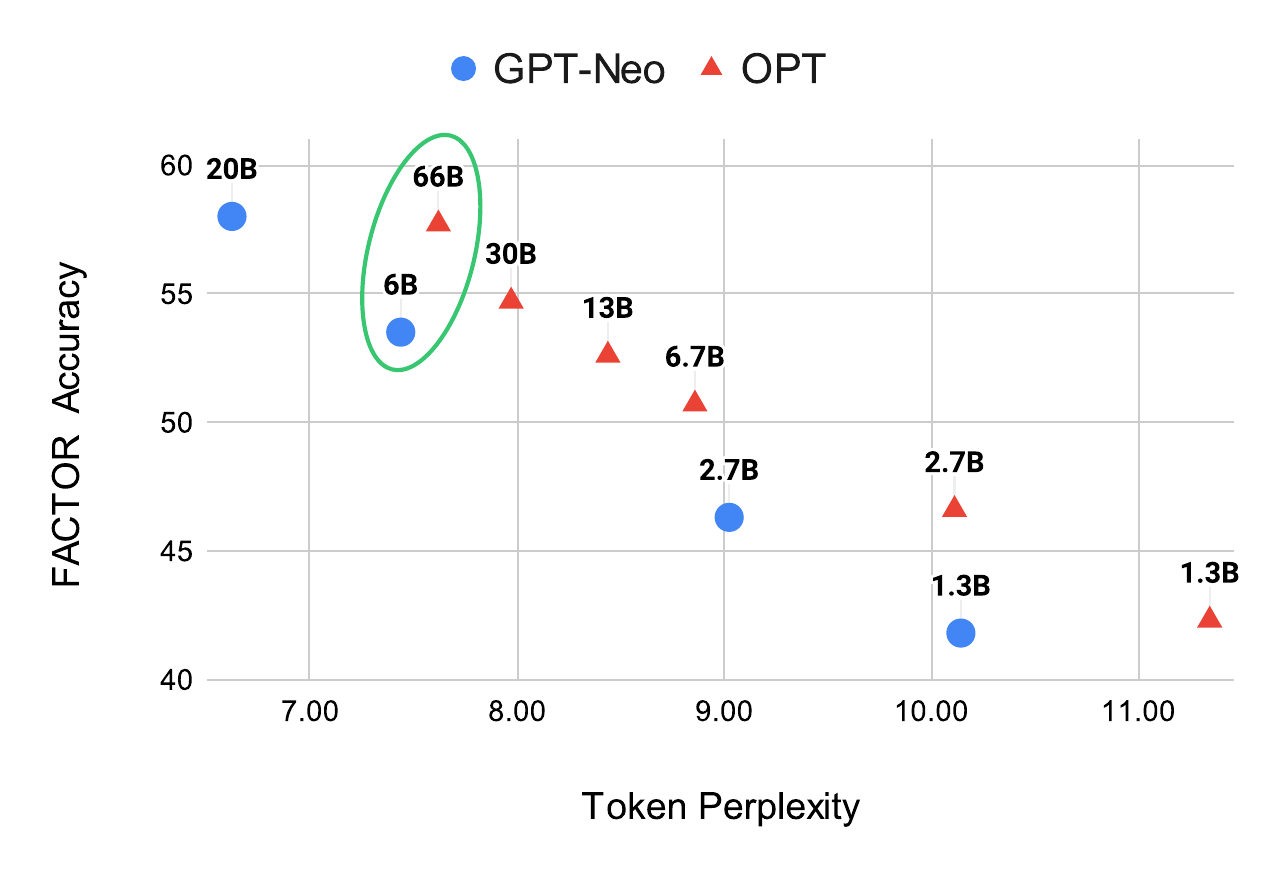}
\vspace{-15pt}
\caption{\wikimethod scores versus LM perplexity on Wikipedia for LMs from the GPT-Neo model family (blue circle, sizes 1.3B-20B) and the OPT model family (red triangle, 1.3B-66B). Labels indicate sizes (in billions). The two may disagree on ranking, \textit{e.g.}, the OPT-66B LM has higher perplexity but better \wikimethod accuracy than the GPT-J-6B LM (marked in green circle). In \S\ref{sec:generation_exeriment} we annotate text generated out of both models and show that better \wikimethod is predictive of more factual text generation.}
\label{fig:ppl_blowup}
\end{figure}

We further show that \method accuracy and LM perplexity are correlted but can sometime induce different orderings between LMs (\S\ref{sec:perplexity}). This highlights that \method and perplexity capture different aspects of the LMs' performance (see Figure~\ref{fig:ppl_blowup}). In \S\ref{sec:generation_exeriment}, we report findings of a manual annotation effort over $1,200$ generated completions, which reinforces \method accuracy as predictive of factuality in open-ended generation.

\section{Related Work}\label{sec:2}

\paragraph{Factuality Evaluation} 

The subject of factuality evaluation has been extensively studied in downstream tasks such as summarization, fact-verification and dialog \cite{honovich-etal-2022-true, huang2021factual, chen-etal-2021-factuality-checkers, tam2022evaluating}. These works typically focus on \textit{factual consistency}, evaluating whether a generated text is supported by a reference text or context (\textit{e.g.}, source document and generated summary). 

Another popular approach suggests probing LMs' internal factual knowledge by using slot filling tasks, \textit{e.g.}, ``Barack Obama was born is [MASK]'' ~\citep{petroni-etal-2019-language,petroni-etal-2021-kilt,roberts-etal-2020-much,jiang-etal-2020-know,elazar-etal-2021-measuring, li-etal-2022-pre, zhong-etal-2021-factual, peng-etal-2022-copen, mallen-etal-2023-trust}. These works test LMs in a simplified, synthetic setting. 
 
\method differs from the above methods as it aims at evaluating factuality in a natural open-ended text generation setting.
In such setting, the context may be needed to reason over the evaluated factual statement, while the factual statement may not be evident in the context (unlike summarization). 

Recent works proposed scoring the factuality of free-form LM generations samples \cite{min2023factscore, lee2022factuality}. However, these approaches lack control over the evaluated facts and are biased towards common facts generated by the LM.

\paragraph{Contrastive Datasets} Contrastive evaluation, in which a model is tested to discern between similar positive and negative examples, is widely used in various tasks \cite{sennrich-2017-grammatical, burlot-yvon-2017-evaluating, glockner-etal-2018-breaking, kaushik2019learning}. For factuality evaluation, negative examples are obtained by perturbing factual claims.
This is done through human annotation, rule-based or model based heuristics \cite{schuster-etal-2021-get, liu-etal-2022-token, gupta-etal-2022-dialfact}. Following recent works on benchmarks generation \cite{perez2022discovering}, we employed Instruct-GPT to generate non-factual claims, as described in the following section.
\begin{table*}[t!]
\small
\centering
\begin{tabular}{ll}
\toprule

\multirow{3}{105pt}{\textbf{Original text}\\  (completion in bold)} & \textit{...In 1982, Donne was appointed as the first Queen's Representative} \\
& \textit{to the Cook Islands. \textbf{After completing his term, he became Chief}} \\
 & \textit{\textbf{Justice of Nauru and Tuvalu in 1985.}} \\
\midrule

\midrule
\textbf{Error Type} & \textbf{Example} \\
\midrule

\multirow{2}{50pt}{\textbf{Entity}} & \textit{After completing his term, he became \textcolor{red}{the Queen's Representative to}} \\
 & \textit{\textcolor{red}{the Cook Islands} in 1985.} \\
\midrule
\multirow{2}{50pt}{\textbf{Predicate}} & \textit{After completing his term, he \textcolor{red}{declined the position of} Chief Justice} \\
 & \textit{of Nauru and Tuvalu in 1985.} \\
\midrule
\multirow{2}{80pt}{\textbf{Circumstance}} & \textit{After completing his term, he became Chief Justice of Nauru and} \\
 & \textit{Tuvalu in \textcolor{red}{1987}.} \\
\midrule
\multirow{2}{50pt}{\textbf{Coreference}} & \textit{After completing \textcolor{red}{her} term, \textcolor{red}{she} became Chief Justice of Nauru and} \\
 & \textit{Tuvalu in 1985.} \\
\midrule
\multirow{2}{50pt}{\textbf{Link}} & \textit{\textcolor{red}{Before} completing his term, he became Chief Justice of Nauru and} \\
 & \textit{Tuvalu in 1985.} \\
\bottomrule
\end{tabular}
\caption{Error types examples. The original text (top) consists of a prefix and a completion sentence (marked in bold). Each example introduce different perturbation over the original completion of different type (edit marked in red).}
\label{tab:negxamples}
\end{table*}
 
\section{The \method Evaluation Approach}\label{sec:3}

This section outlines our proposed approach: Factual Assessment via Corpus TransfORmation, or \method. Given a corpus, we define a multi-choice task where each example is comprised of a multi-sentence prefix, a single \textit{factual} next sentence completion, and three \textit{non-factual} alternative completions (Figure \ref{fig:intro_fig}). In \S\ref{sec:3.1} we present several properties required of a \method benchmark, and describe the error verticals along which we generate non-factual alternatives. We then explain our \method dataset creation pipeline, which automatically generates a \method benchmark from a given corpus (\S\ref{sec:3.2}). Finally, we apply this pipeline to two corpora Wikipedia and news, and a long-form question answering dataset, creating \wikimethod, \newsmethod and \qamethod. We verify the quality of these datasets through manual annotations against the required properties (\S\ref{sec:3.3}).

\subsection{The Evaluation Task: \textit{\method}}\label{sec:3.1}

We describe the \method multi-choice factual evaluation task. Each example of our task contains a prefix text $t$, along with four possible full sentence completions, of which only one is factually correct. We choose the original completion (\textit{i.e.}, the continuation of $t$ in the corpus) as the factually correct one. The correct completion is denoted as $c^+$, and the non-factual completions as $\mathcal{C}^-=\{c^-_1, c^-_2, c^-_3\}$. We evaluate models by measuring the percentage of examples where they assign the highest mean log-probability to $c^+$. Formally, a model is correct on a given example if:

\begin{equation}
    c^+=\argmax_{c\in\{c^+\}\cup \mathcal{C}^-}\frac{\log p(c|t)}{|c|},
\end{equation}
where $|c|$ is the length of completion $c$ in tokens.
We refer to the percentage of correct examples as the \method accuracy.

We require each of the ``incorrect'' completions $c^{-}\in\mathcal{C}^-$ to satisfy the following properties:
\begin{enumerate}
    \item \textit{Non-factuality}: $c^{-}$ contains a false claim;
    \item \textit{Fluency}: $c^{-}$ is grammatical; 
    \item\textit{Similarity to the factual completion}: $c^{-}$ has a small edit-distance from $c^+$.
\end{enumerate}

The second and third properties make it harder to distinguish between the factual and non-factual completions for reasons other than their factual correctness, such as fluency or style.
Furthermore, it is desirable that the non-factual completions be logical and self-consistent, to make them more difficult to eliminate.
For example, modifying $c^+=$\textit{``They got married in 2010 and divorced in {\textbf{2017}}''} by changing \textit{2017} to \textit{2009}, results in a non-factual completion which can be discarded by knowing the temporal relation between marriage and divorce.

\paragraph{Error Types} Non-factual completions in a \method dataset should cover diverse factuality error types. To do so, we adopt the error typology introduced in FRANK \cite{pagnoni-etal-2021-understanding}. While they introduced their error typology to categorize factual inconsistencies of generated summaries w.r.t.\ the source document, we instead leverage this typology to vary the type of factual inconsistencies that hold between non-factual completions and the prefix and completion ($t$ and $c^+$). We focus on the five error types from two error categories: semantic frame and discourse (examples in Table \ref{tab:negxamples}):

\begin{itemize}
  \item \textit{Predicate} error: a predicate that is inconsistent with $c^+$ or $t$.
  \item \textit{Entity} error: The subject or object of a predicate are inconsistent with $c^+$ or $t$.  
  \item \textit{Circumstance} error: The completion contains information describing the circumstance of a predicate (\textit{e.g.}, location, time, manner) that is inconsistent with $c^+$ or $t$.
  \item \textit{Coreference} error: The contradiction is inconsistent with a pronoun/reference in $c^+$ or $t$, referring to a wrong or non-existing entity. 
  \item \textit{Link} error: $c^-$ is inconsistent with $c^+$ or $t$ in the way that different statements are linked together (causal/temporal links). 
\end{itemize}

\subsection{Generating \method Benchmarks}\label{sec:3.2}

Given an evaluation corpus, we generate a \method benchmark automatically. The process is designed to meet the requirements presented in \S\ref{sec:3.1}, and follows a four-stage pipeline: (1) prefix and completion selection, (2)~non-factual completion generation, (3)~non-factual completion filtering, and (4)~non-factual completion selection.

\subsubsection{Prefix and Factual Completion Selection}\label{subsec:3.2.1} We select a single sentence from each document as a factual completion $c^+$. We exclude headlines and sentences with less than 10 words. The prefix $t$ is the entire text preceding $c^+$ in the document. 

\subsubsection{Non-factual Completions Generation}\label{subsec:3.2.2} Given a prefix $t$ and its original completion $c^+$, we use InstructGPT (davinci-003; \citealt{ouyang2022training}) to generate a set of contradictory completions. We designed a specific prompt instructing the model to generate contradictions corresponding to each type of error.\footnote{App.~\ref{app:prompts} lists the full prompts for each error type.} We only apply each prompt to sentences that are relevant to its error type (determined through simple heuristics, see App.~\ref{app:rel-err-type}). The prompts are designed as follows:
\begin{itemize}
  \item Multiple contradiction generation: the model is prompted to generate multiple subsequent contradictions in each sampling operation. 
  Preliminary experiments showed that this sampling practice improves diversity compared to multiple independent completion sampling. 
  \item Edit planning: for each contradiction, the model first explicitly generates the planned edits over the original completion, and then applies those edits by writing the entire \textit{modified} completion (similar to chain-of-thought prompting;~\citealt{wei2022chain}). 
  For instance, the coreference error in Table~\ref{tab:negxamples} is generated by explicitly writing the edits ("Changes: `his' to `her'") and then the contradiction. This encourages the model to make minimal edits. 
\end{itemize}

\subsubsection{Non-factual Completions Filtering}\label{sec:completion_filtering}\label{subsec:3.2.3} 
We considered the set of generated completions as candidates for non-factual completions. 
We applied automatic tools to filter out (i) \textit{non-contradictory} and (ii) \textit{non-fluent} completions.

\paragraph{Non-Contradictory Completions} Given a candidate completion $c$, we assert that it is indeed contradictory to the original completion $c^+$ by applying an NLI model.\footnote{We used DeBERTa-large model \cite{he2020deberta} fine-tuned on the MNLI dataset \cite{williams-etal-2018-broad} from Hugging Face: \href{https://huggingface.co/microsoft/deberta-large-mnli}{microsoft/deberta-large-mnli}.} The \textit{premise} is set to be $c^+$ along with its near context (\textit{i.e.}, the last tokens of the prefix $t$; denoted by $t_\text{near}$). 
The \textit{hypothesis} is set to be $c$, also preceded by $t_\text{near}$.
 We selected generations classified as contradictory by the NLI model with a probability higher than $\tau_{\textrm{NLI}}$, \textit{i.e.}: 
 \begin{equation*}\label{eq:t_nli}
p_{\textrm{NLI}}(\text{contradiction}\ |\  [t_{\textrm{near}};c^+], [t_{\textrm{near}};c])) > \tau_{\textrm{NLI}}
\end{equation*}
We chose $\tau_{\textrm{NLI}}=0.6$ (except for contradictions generated by the coreference error prompt, where we set $\tau_{\textrm{NLI}}=0.3$) after using a manual validation process detailed App.~\ref{app:filters-thr}. 

\paragraph{Non-Fluent Completions} 
To verify that $c$ is a fluent completion we use GPT2-Small 
\cite{gpt2} scores, similar to \citet{gupta-etal-2022-dialfact}: We filter out generations with mean log-likelihood lower than the original completion's by a fixed margin $\tau_{\textrm{LM}}$. Using a manual validation, we set $\tau_{\textrm{LM}}=0.2$ (see App.~\ref{app:filters-thr}). Formally, we selected a completion $c$ if it satisfies:
\begin{equation*}\label{eq:t_lm}
\frac{\log p(c)}{|c|} > \frac{\log p(c^+)}{|c^+|} - \tau_{\textrm{LM}}
\end{equation*}

\subsubsection{Non-factual Completion Selection}\label{subsec:3.2.4} 
Finally, we select non-factual completions $c^-_1,c^-_2, c^-_3$ from the filtered candidates. 
For increased error type diversity, we choose one completion per type, and repeat types only when not enough generations meet the \S\ref{sec:completion_filtering}'s criteria.

\begin{figure*}[!htb]
 \begin{subfigure}{0.32\textwidth}
  \includegraphics[width=\linewidth]
  {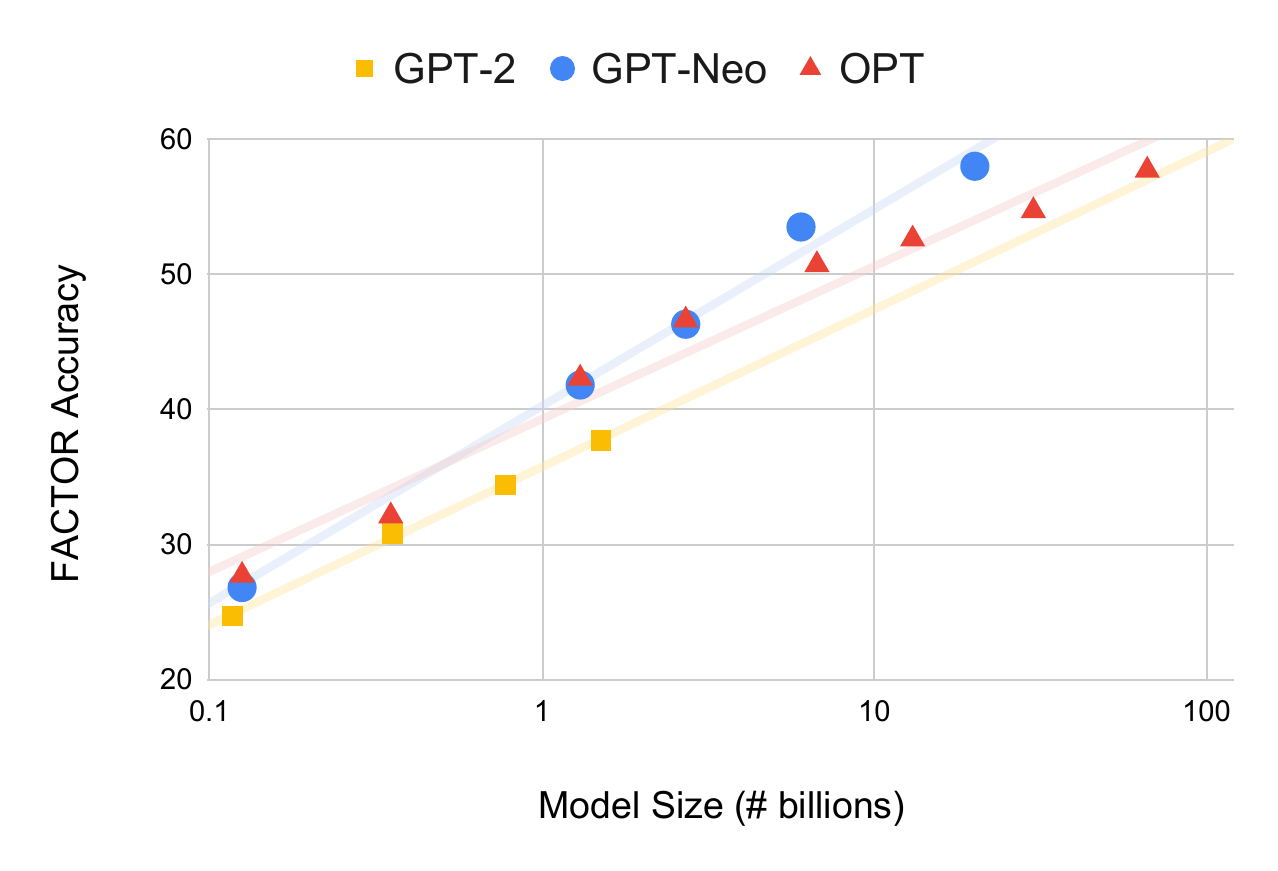}
  \caption{\wikimethod}\label{fig:sub1}
\end{subfigure}\hfill
\begin{subfigure}{0.32\textwidth}
  \includegraphics[width=\linewidth]
  {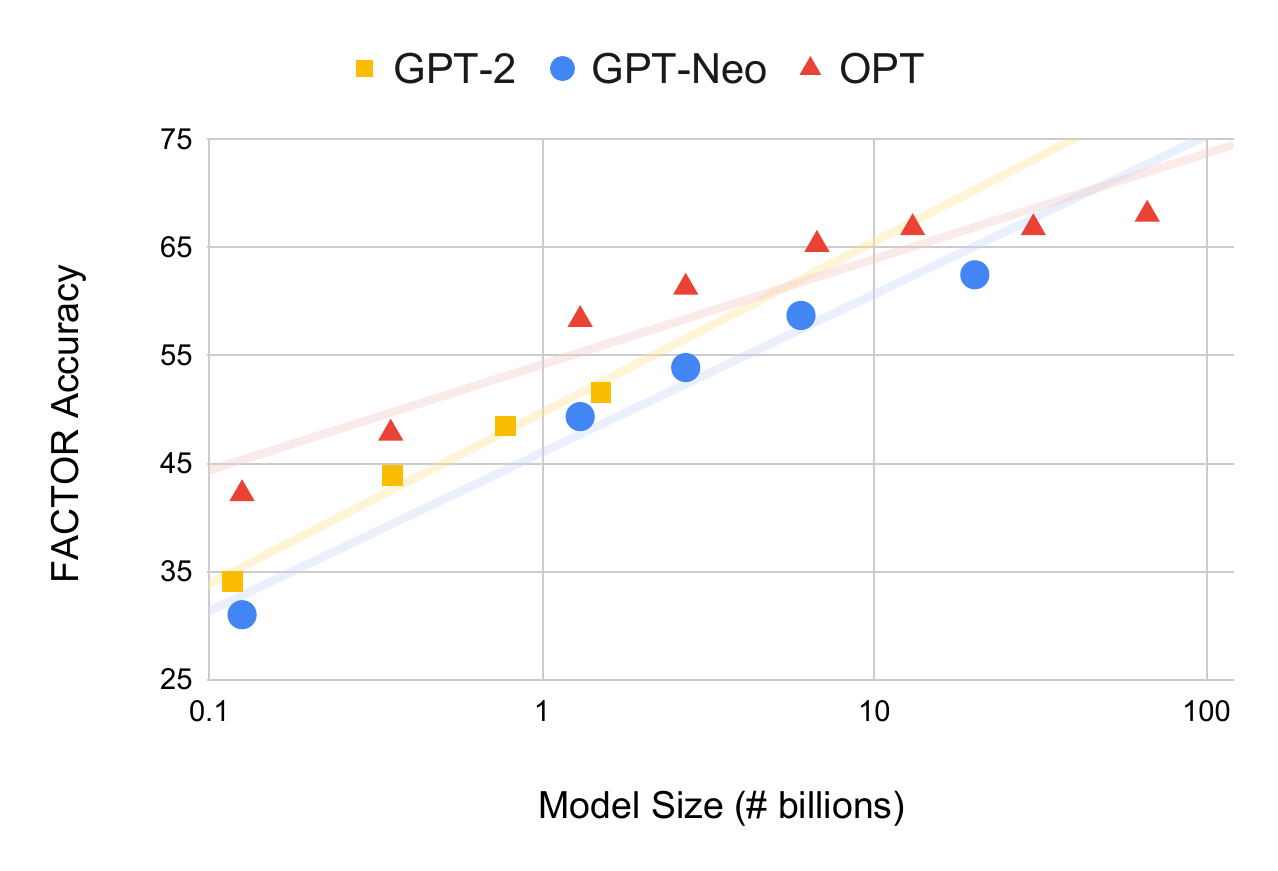}
  \caption{\newsmethod}\label{fig:sub2}
\end{subfigure}\hfill
\begin{subfigure}{0.32\textwidth}%
  \includegraphics[width=\linewidth]{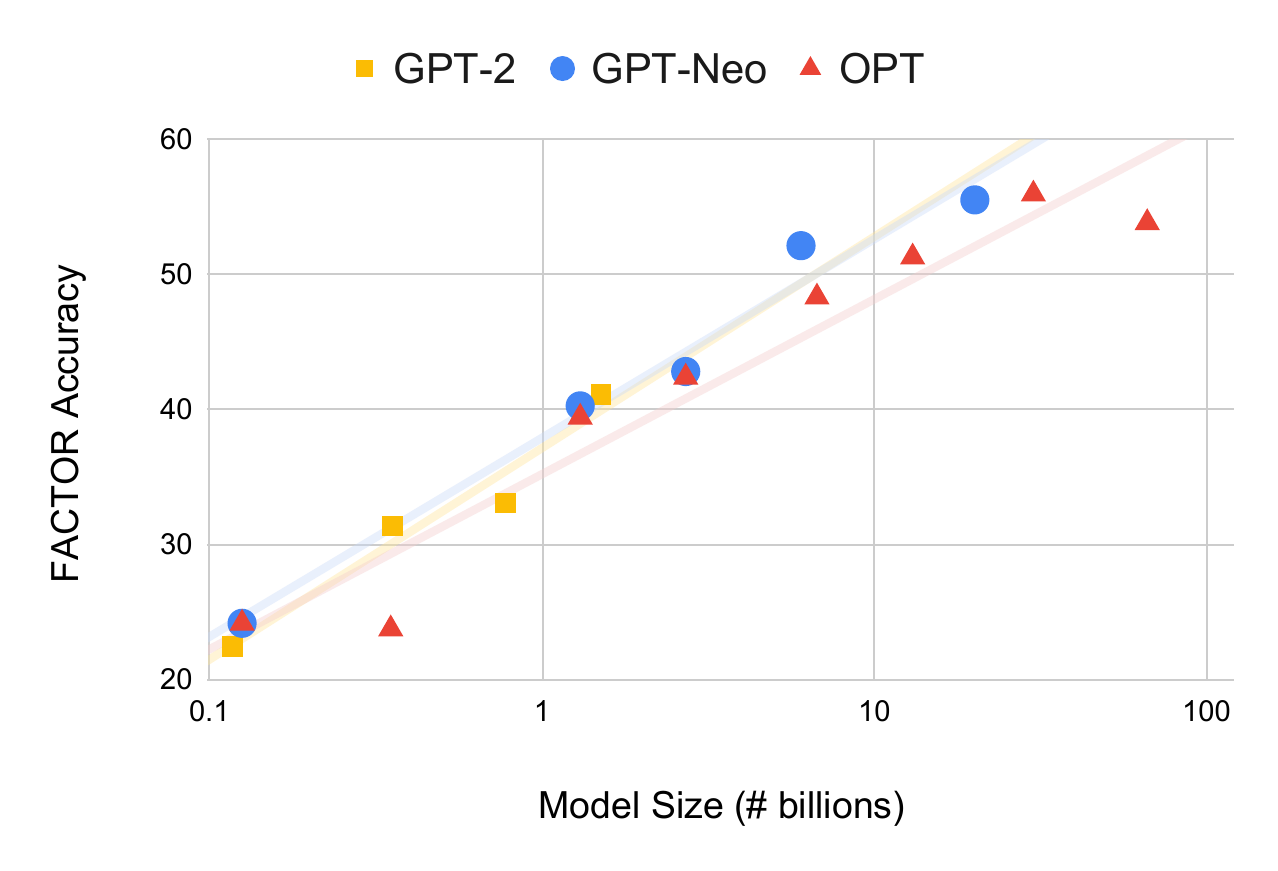}
  \caption{\qamethod}\label{fig:sub3}
\end{subfigure}
\caption{Accuracy per model size for \wikimethod (left), \newsmethod (center), and \qamethod (right) for models from GPT-2 (yellow square), GPT-Neo (blue circle), and OPT (red triangle) families.}
\label{fig:fc_res}
\end{figure*}

  
  

\begin{table}[t!]
\small
\centering
\begin{tabular}{lccc}
\toprule
\textbf{Property} & \textbf{Wiki} & \textbf{News} & \textbf{Expert}\\
\midrule
 Non-factual & 97.6 & 98.3 & 97.5 \\
 Fluent & 94.0 & 97.0 & 96.7 \\
 Self-Consistent & 87.4 & 87.3 & 83.8 \\
 \midrule
 Edit-Distance & 2.3$\pm$(1.4)& 2.1$\pm$(1.4)& 4.0$\pm$(3.1)\\
\bottomrule
\end{tabular}
\caption{Validation results: percentage of generation that meet each desired property, estimated by manual annotation over sub-samples (top), and mean edit-distance between the generations and their factual completion (bottom).  
}
\label{tab:data-stats}
\end{table}

\subsection{Applying \method to Knowledge Intensive Domains}\label{sec:3.3}
We focused on three knowledge intensive domains: Wikipedia (encyclopedic knowledge), news (current events) and long-form question answering in specific domains. We constructed the following evaluation datasets:  
\begin{itemize}
  \item \textit{\wikimethod:} based on the Wikipedia section of The Pile's validation split \cite{pile}, containing $2994$ examples.
  \item \textit{\newsmethod:} based on Reuters articles published after $1/10/2021$, extracted from The RefinedWeb Dataset \cite{penedo2023refinedweb}. The dataset consists of $1036$ examples.
  \item \textit{\qamethod:} based on the validation and test splits of ExpertQA \cite{malaviya23expertqa}, a long-form expert-curated question answering dataset spanning various fields, which suits the motivation of \method to evaluate rare facts. Each document in the corpus is a concatenation of a question-answer pair. The dataset consists of $236$ examples.  
\end{itemize}

\subsubsection{Dataset Validation}

To validate that our \method benchmarks meet the required properties detailed in~\S\ref{sec:3.1}, we manually evaluated a sub-sample from each dataset. We sampled $138$ examples from \wikimethod, $100$ examples from \newsmethod and $80$ examples from \qamethod, containing $414$, $300$ and $240$ generations overall. Each generation was annotated w.r.t.\ the properties manifested in \S\ref{sec:3.1}, namely whether they were (1)~non-factual, (2)~fluent, and (3)~self-consistent. To assess datasets diversity, we annotated the contradictions in accordance with the error typology of~\citet{pagnoni-etal-2021-understanding}, described in \S\ref{sec:3.1}. 
We verified that the non-factual completions are minimally edits variants of the factual completion by measuring mean edit distances. 

Validation results in Table \ref{tab:data-stats} show that for all datasets, almost every generated completion indeed contradicts the original one, was fluent, and was self consistent. Table \ref{tab:data-dist} shows the error type distribution, indicating that \method yields diverse contradiction types. Semantic frame errors (Entity, Predicate, and Circumstance) were more prevalent than discourse errors (Link and Coreference), as more sentences are suited for these type of errors. 

\begin{table}[t!]
\small
\centering
\begin{tabular}{lccc}
\toprule
\textbf{Type} & \textbf{Wiki} & \textbf{News} & \textbf{Expert}\\
\midrule

Predicate & 25.4 & 31.3 & 47.1\\
Entity & 42.8 & 48.0 & 38.8 \\
Circumstance~~~~~~~~ & ~~~24.2~~~ & ~~~16.0~~~ & ~~7.1 \\
Coreference & ~~4.4 & ~~2.3 & ~~2.9\\
Link & ~~3.2 & ~~2.3 & ~~4.2\\
\bottomrule
\end{tabular}
\caption{Annotated error type distribution for \wikimethod (Wiki), \newsmethod (News), \qamethod (Expert).
}
\label{tab:data-dist}
\end{table}

\section{Experimental Setup}\label{sec:4}
We used FACTOR benchmarks to evaluate factual knowledge of LLMs across varying model families. We describe the experimental setup below.

\subsection{Datasets}\label{sec:datasets} 
The \wikimethod, \newsmethod and \qamethod datasets are described in \S\ref{sec:3.3}. For perplexity evaluation (\S\ref{sec:perplexity}), we selected a subset of $300$ Wikipedia articles from the documents \wikimethod is based on (${\sim}{367}$K tokens).

\subsection{Models}\label{sec:models} 
We performed our experiments over a set of open source models: four models of GPT-2 family (110M–1.5B; ~\citealt{gpt2}), five models from the GPT-Neo family (125M–20B;~\citealt{gpt-neo,black-etal-2022-gpt,gpt-j}), and eight models of OPT (125M–66B;~\citealt{zhang2022opt}). We capped the sequence length at $1024$ tokens to compare all models directly.

The corpora that our \method benchmarks were constructed from were not used for training any of the examined models. \newsmethod is based on articles published after 1/10/2021, while \qamethod is based on examples written in 2023. Both are beyond the models' data cutoff date. \wikimethod is based on Wikipedia documents from The Pile's validation split, which is not part in any of the models' training sets.
(OPT and GPT-Neo models were trained on The Pile's training split, GPT-2 models were not trained on Wikipedia).

\subsection{Retrieval-Augmented Models}\label{sec:ralm} 
In \S\ref{sec:results_retrieval}, we present evaluations of retrieval-augmented variants of the models. To that end, we adopted the In-Context RALM (IC-RALM) framework of \citet{ram2023context}, where the retrieved document is prepended to the LLM's input, without any further training or specialized LLM architecture. In IC-RALM, 
a retriever is called every $s$ tokens (\textit{i.e.}, the \textit{stride}), with a query comprised of the last $\ell$ tokens. The LLM is run with the concatenated input to assign log-probabilities to the next $s$ tokens. We used the lexical BM25 \cite{bm25} over Wikipedia corpus,\footnote{We used the Wikipedia corpus of \citet{karpukhin-etal-2020-dense}, based on the dump from Dec. $20, 2018$.} excluding the evaluated docs; and set $s=8$, $\ell=32$.
\section{Factual Knowledge Evaluation Results}

This section describes the experimental evaluation of LLM factuality using our \method benchmarks. In \S\ref{sec:results_size} we show that \method accuracy increases with model size but also depends on the training data (different model families differ in scores). In \S\ref{sec:results_retrieval}, we show that retrieval augmentation of the LM improves \method accuracy, positioning it as the first automatic measure of factuality improvement for retrieval augmented LMs. Finally, in \S\ref{sec:perplexity}, we show that the pairwise model ranking of corpus perplexity and \method accuracy can differ significantly. This outcome, along with manual validation of the correlation between \method accuracy and factual generation in \S\ref{sec:generation_exeriment}, solidifies \method accuracy as a novel automatic measure for evaluating the proneness of an LM to generate factual information in a certain domain.
 
\begin{figure*}[t!]
\centering
\subfloat{%
  \includegraphics[clip,width=0.625\textwidth, height=125pt]{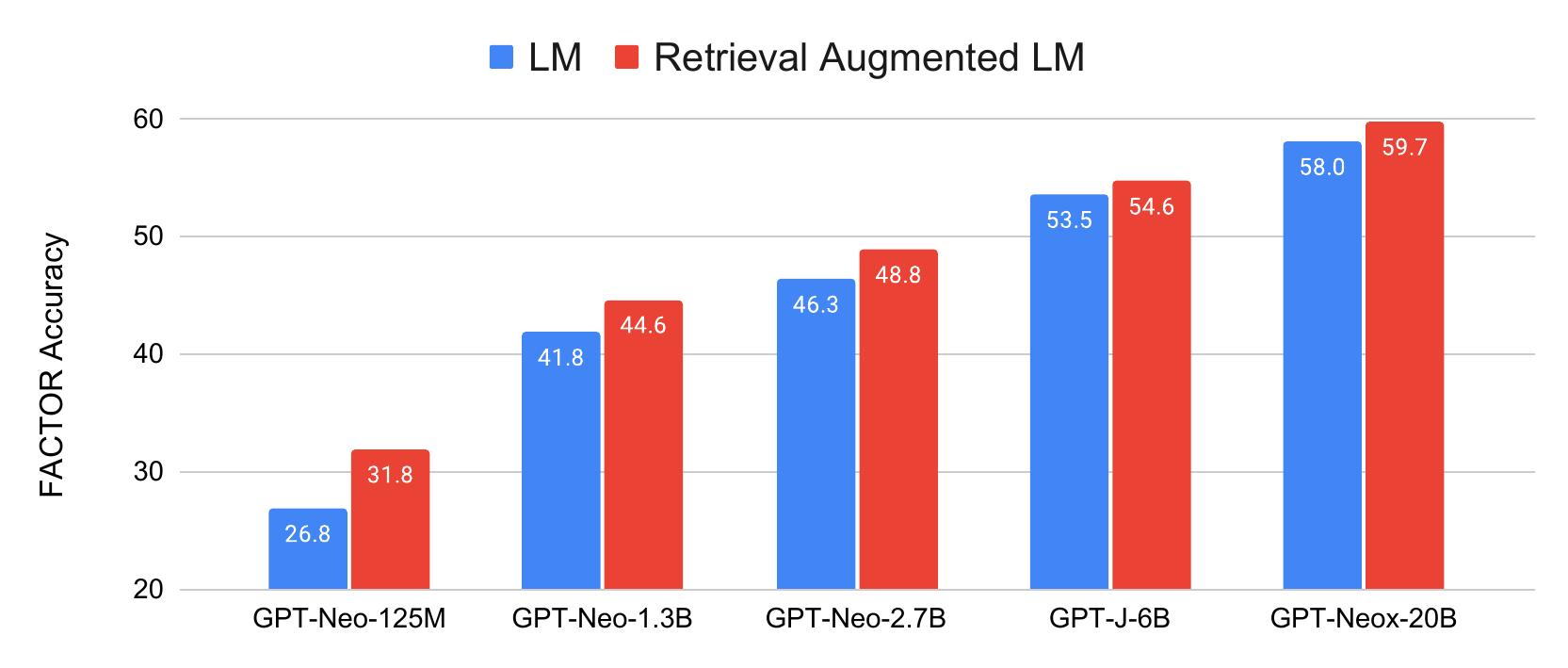}%
}

\subfloat{%
  \includegraphics[clip,width=0.9\textwidth, height=125pt]{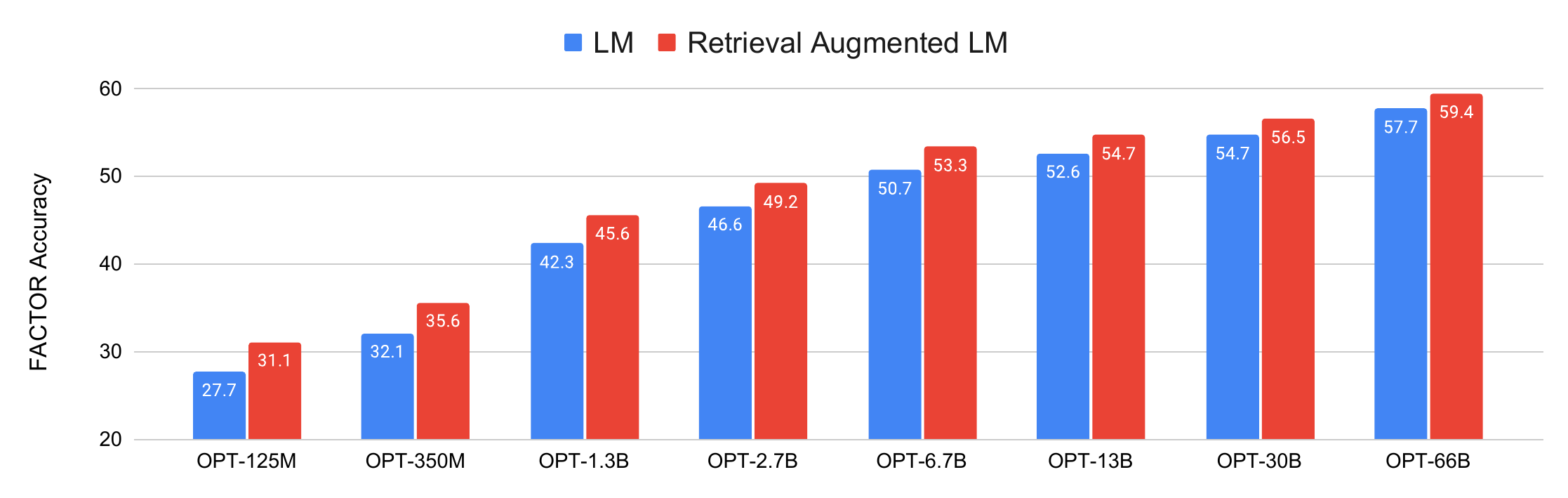}%
}
\caption{Factual accuracy over \wikimethod for GPT-Neo and OPT models, compared to their IC-RALM variants. IC-RALM leads to consistent improvement for all models. 
}
\label{fig:ralm_results}
\end{figure*}


\subsection{Factual Knowledge Improves with Model Size}\label{sec:results_size}

We evaluate GPT-2, GPT-Neo, and OPT models 
on \wikimethod, \newsmethod and \qamethod (Figure~\ref{fig:fc_res}). 
Larger models generally outperform smaller ones within the same model family. However, even the largest models are capped at $58.0$\% (GPT-NeoX-20B), $68.1$\% (OPT-66B) and $55.9$\% (OPT-30B)  on \wikimethod, \newsmethod and \qamethod respectively, indicating the benchmarks are challenging. Recent works \cite{chuang2023dola, kai2024sh2} use \wikimethod and \newsmethod to evaluate models from the LLaMA family \cite{touvron2023llama} and show similar trends.

We observe that all models achieve higher \method accuracy on news comparing to the other two domains. This may be because news articles cover specific events, making the prefix more useful for detecting factual completions (further discussion in App.~\ref{app:generation_exp}). When comparing different model-families, we find that the OPT models leads on \newsmethod, while the GPT-Neo family leads on \wikimethod. This implies that the different data sources used for training these two model families are suited to different domains.

\subsection{The Effect of Retrieval Augmentation on Factual Knowledge }\label{sec:results_retrieval}

Next, we ask: \textit{Can \method accuracy be improved by augmenting models with a retrieval component?} 
Importantly, while a clear motivation for retrieval augmentation is factual grounding of LMs, no existing metrics allow direct measurement of it in a text generation setting.
We propose \method accuracy as an alternative to the course measure of LM perplexity, which is often used to assess these methods~\cite{knn-lm,retro,ram2023context,shi2023replug}.

We compared the \method accuracy of LLMs to that of their retrieval-augmented counterparts, implemented following the IC-RALM framework (\S\ref{sec:ralm}; \citealt{ram2023context}). Figure~\ref{fig:ralm_results} show the results for GPT-Neo and OPT \wikimethod. 
We observed consistent gains from augmenting the models with retrieval. 
These results highlight that grounding the model in an external corpus can improve its factuality. 
Since the retriever used in our experiments is used in an ``off-the-shelf'' manner, we speculate that further performance boosts may be gained by a retriever system specialized for this task 
\citep{izacard2022atlas,ram2023context}. 

 Another interesting finding is that the \textit{relative} gains in \method accuracy obtained by IC-RALM, are more moderate compared to the relative gains in perplexity over WikiText-103 \citep{wikitext}, reported by \citet{ram2023context}.
 We explore the connection between the two in the next section. 

\subsection{Perplexity Correlates but is not Always Aligned with \method Accuracy}\label{sec:perplexity}

We investigate whether \method accuracy adds additional information beyond perplexity, when used as a comparative metric for selecting which LM to use within a certain corpus. Figure~\ref{fig:ppl_blowup} shows the \method accuracy of models on Wiki-\method, compared to their token-level perplexity on the Wikipedia section of The Pile's validation set (\S\ref{sec:datasets}) (App.~\ref{app:full-ppl} includes all evaluated models).
Overall, we observe a high correlation between the two metrics. However, there are cases where they disagree (\textit{i.e.}, a pair of models where one is better when measured by perplexity but worse in terms of \method accuracy). For example, GPT-Neo-2.7B is significantly better than OPT-2.7B in terms of perplexity ($9.0$ vs. $10.1$), but slightly worse in terms of \method accuracy  ($46.3$\% vs. $46.6$\%).
In addition, GPT-J-6B has lower perplexity compared to OPT-66B ($7.4$ vs. $7.6$), while OPT-66B is significantly better in terms of \method accuracy ($57.7$\% vs. $53.5$\%). 
This finding suggests that (i) \method accuracy offers a complementary view of models' performance, not necessarily captured by perplexity, and (ii) improvements in perplexity do not necessarily imply better factuality.

\section{Factuality in Open-Ended Generation}\label{sec:generation_exeriment}
This section explores the connection between \method accuracy and factuality in open-ended generation, via human annotations.

\subsection{Experimental Setup}\label{subsec:generation_setup}

We selected tuples of prefix, original completion and non-factual completion $(t, c^+, c^-)$ from \wikimethod. 
We then manually identified the \textit{minimal factual claim} modified by $c^-$, denoted by $f$. For example, the predicate error from Table~\ref{tab:negxamples}, in which ``\textit{became}'' was replaced with \textit{``declined the position of''}, the edit relates to the minimal fact ``\textit{Donne became Chief Justice of Nauru and Tuvalu}''. 

We let LLMs generate free text, conditioned on the prefix and the completion until the edit induced by $c^-$. 
Formally, let $c$ be the common prefix of $c^+$ and $c^-$ (in the predicate error example, $c$ is ``\textit{After completing his term, he}"). The LLM is conditioned on the concatenation of $t$ and $c$. 
The LLM might generate the correct fact, text violating it, or other completion that does not refer to it.
For each example we manually annotated whether the generated text is \textit{true},  \textit{false}, or \textit{neutral} w.r.t.\ $f$. 

We analyzed two models with a similar token-level perplexity but a significant gap in \method accuracy: GPT-J 6B and OPT-66B (marked in a green circle in Figure \ref{fig:ppl_blowup}). For each model, we considered two groups of examples: examples with $c^{+},c^{-}$ pairs for which the model was \textit{right}, \textit{i.e.}, the model assigns larger mean log-likelihood to $c^{+}$ compared to $c^{-}$, and pairs for which the model was \textit{wrong} (the complement set). We sampled three generations per example for $100$ examples from each group and for each model. Overall, we created $1200$ generations. We filtered some of the samples due to ill-formatted generations or non-contradictory completions ($14.5$\% of all samples).

\begin{table}[t!]
\small
\centering
\begin{tabular}{llc}
\toprule
\textbf{Model} & \textbf{Subset} & \textbf{Fact. Accuracy} \\
\midrule
\multirow{3.5}{50pt}{\textbf{GPT-J 6B}}
& Right & 30.0\% \\
& Wrong & 10.5\% \\
\cmidrule{2-3}
& All (Weighted) & 24.8\% \\
\midrule
\multirow{3.5}{50pt}{\textbf{OPT-66B}}
& Right & 46.6\% \\
& Wrong & ~~4.6\% \\
\cmidrule{2-3}
& All (Weighted) & \textbf{38.8}\% \\
\bottomrule
\end{tabular}
\caption{Manual factuality annotation results for OPT-66B and GPT-J 6B. For each model, we present the results per \textit{right} and \textit{wrong} subsets. Bottom row shows the weighted average between the \textit{right} and \textit{wrong} variants w.r.t to the \textit{right}/\textit{wrong} pairs of \wikimethod.
}
\label{tab:factuality_generation}
\end{table}

\subsection{Results}\label{subsec:generation_setup}

We assess model's knowledge of the minimal facts through manual annotation. We only considered relevant generations for their minimal fact $f$, excluding "neutral" generations (59.5\% and 54.3\% for GPT-J 6B and OPT-66B, respectively). For each model, we measure the percentage of generated texts that are true w.r.t. $f$ in the "right" and "wrong" subsets separately. We obtained the overall \method accuracy by weighting the subsets results according to their distribution in \wikimethod. 
Results in Table~\ref{tab:factuality_generation} (full results in App.~\ref{app:generation_exp}).

\paragraph{Accuracy over \wikimethod is linked with factuality in open-ended generation.} For cases where models were \textit{wrong}, they generated more false claims regarding their minimal fact. For example, OPT-66B only generated a true claim $4.6$\% of the times it was wrong, compared to $46.6$\% for when it was right. This suggests that \method accuracy can shed light on the model's ability to generate factual claims accurately.

\paragraph{As a comparative metric, accuracy over \wikimethod aligns with factuality in open-ended generation.} There were gaps in factuality annotation between OPT-66B and GPT-J 6B: OPT-66B generated \textit{true} claims $38.8$\% of the time, while GPT-J 6B generated only $24.8$\%. This aligns with the models' performance over \wikimethod, despite sharing similar perplexity on Wiki. This suggests that \method is a better proxy for measuring model factuality in a specific domain.

\section{Discussion}

This paper introduces \method, a novel way to evaluate LMs' factuality. \method creates an evaluation benchmark from a corpus, consisting of factual statements and non-factual variations. 
By comparing the LM's likelihood of factual claims with non-factual variants, \method score captures the LM's propensity to generate factual information.

Metrics for measuring factual knowledge over a given corpus are lacking. Prior works used perplexity, which may be affected by factors other than factual knowledge and does not contrast facts with false statements. \method focuses the language modeling task on factuality by taking a contrastive approach. Our experiments show that \method ranks models differently than perplexity and is more aligned with factuality in open-ended generation. These findings highlight the importance of negative examples for evaluating factuality. Moreover, they indicate that incorporating negative examples into training sets might also help optimizing models to be more factual. We leave investigation of training with FACTOR style data to future work.    

Our work joins recent studies on factuality evaluation in a text-generation setting, which proposed to evaluate models by fact-checking the model's generations \cite{lee2022factuality,min2023factscore}. As \method focuses on evaluation over a controlled set of facts, we see these two approaches as complementary; together, they yield a more holistic assessment of LM factuality.

\section*{Limitations}

We point to several limitations of our work. First, since \method benchmarks are generated in an automated way, they may not fully comply with the requirements we define in \S\ref{sec:3.1}, as analyzed in \S\ref{sec:3.3}.   
Second, generating \method benchmarks for different domains may pose new challenges. For instance, the selection of factual completions is straightforward in knowledge-intensive domains, where nearly every sentence in the corpus contains factual information. However, in general cases, a more intricate approach is needed to identify such sentences. Moreover, the generation of non-factual completions is based on a prompted model, specifically designed for the Wikipedia domain. While we observed those prompts applied well for the news domain, their effectiveness may vary in other, more specific domains.   

\section*{Ethics Statement}

Language models' tendency to generate factually inaccurate text raises significant issues. \method allows automatic evaluation of factuality, which can be used to efficiently measure and develop methods for mitigating these risks. However, we stress that when deploying such models in sensitive settings, automatic evaluations may not be sufficient, and human evaluation is required.

\bibliography{anthology,custom}
\bibliographystyle{acl_natbib}
\appendix

\section{Technical Details of \method Data Pipeline}\label{app:pipeline-tech}

\subsection{Identifying Sentences' Relevant Error Types}\label{app:rel-err-type}
For each sentence, we identify the types of edits we can apply to it. First, we use a part-of-speech tagger to detect relevance for entity error (detecting nouns), predicate error (detecting verbs) and coreference error (detecting pronouns). For circumstances errors, we use Named-Entity Recognition taggers to identify sentences containing locations, dates, and time entities. Finally, we search for temporal/causal link words from a predefined set of words, which implies relevance for link errors.

\subsection{Setting Filters Thresholds}\label{app:filters-thr}
As discussed in \S\ref{sec:completion_filtering}, we applied two filters to ensure the quality of the potential completions--an NLI filter (to filter out non-contradictory completions) and an LM filter (to filter out non-fluent completions). To choose the thresholds  $\tau_{\textrm{NLI}}$ and $\tau_{\textrm{LM}}$, we manually annotated 40 samples w.r.t to the properties specified in~\S\ref{sec:3.1} (\textit{i.e.}, (1) contradictory and (2) fluent and self-consistent). We have tested thresholds 0.1-0.9, and chose the threshold which achieved highest precision without filtering out too many samples (max 35\% of the samples). 
For the NLI filter we used DeBERTa-largs model fine-tuned on the MNLI dataset. Best threshold was $\tau_{\textrm{NLI}}=0.6$, with precision of 0.96. Manually evaluating the different contradiction types we have noticed this threshold was too harsh for corefrence contradiction (87.5\% of the completions were filtered out. Therefore we reduced its threshold to 0.3 which filtered out 75\% of the samples). For the LM filter we used GPT2-Small. Best threshold was $\tau_{\textrm{LM}}=0.2$, with precision of 0.78. 
\section{Extended Results and Discussion}
\subsection{Comparison between Perplexity and \method Accuracy over Wikipedia}\label{app:full-ppl}
\begin{figure}[t]
\centering
\hspace*{-10pt}
\includegraphics[width=1.05\columnwidth]{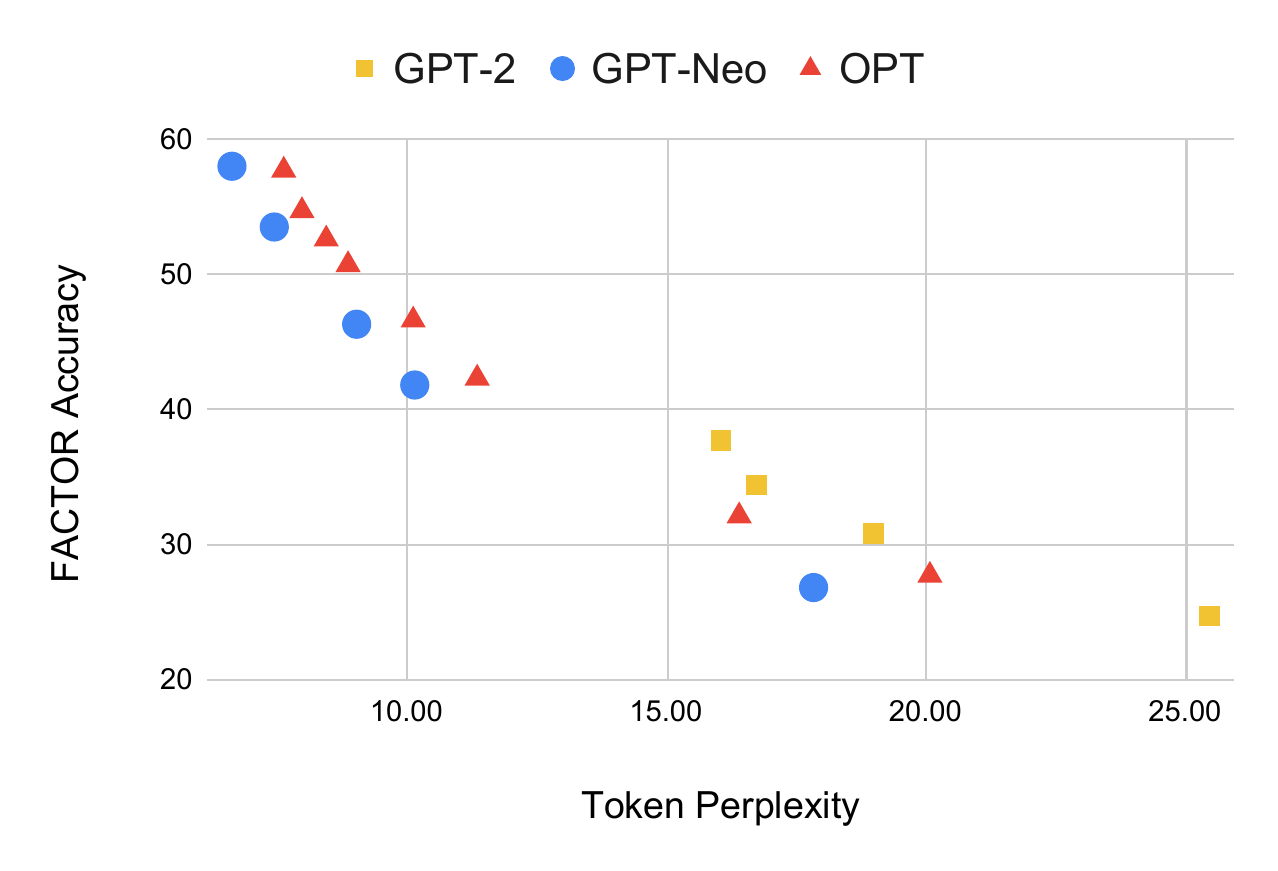}
\vspace{-15pt}
\caption{Accuracy per token perplexity over \wikimethod.}
\label{fig:full_ppl}
\end{figure}
Figure \ref{fig:full_ppl} presents \wikimethod scores versus LM perplexity on Wikipedia. The figure extends Figure \ref{fig:ppl_blowup}, presenting all evaluated LMs: models from the GPT-Neo family (blue circle), OPT family (red triangle) and GPT2 family (yellow square). 
\subsection{Factuality in Open-ended Generation}\label{app:generation_exp}
Table \ref{tab:factuality_generation_app} shows the extended results for the manual factuality annotation for open-ended generation experiment \S\ref{sec:generation_exeriment}. In addition to the overall results, we include the distribution of Neutral/True/False annotations. Notably, most generations are neutral for both models. This highlights the limitation of sampled-based approach for assessing model's factual knowledge.

\subsection{Knowledge of Unseen Facts}\label{app:unseen_facts}
As seen in Figure~\ref{fig:fc_res} in \S\ref{sec:results_size}, 
FACTOR-accuracy is often way above the random baseline of 25\%, indicating that some models succeed in predicting unseen facts. It is possible that the knowledge of these facts is derived from another document in the training data (for example, Wikipedia contains many different articles related to each other, sharing similar factual statements). Another possibility is that an unseen fact is implied by the prefix. We hypothesize that this leads to higher FACTOR scores in the news domain, which often covers specific events, making the prefix more useful for detecting factual completions. Analysis of these cases is non-trivial, and is left for future work.
\section{Dataset Licenses}\label{app:license}

Table \ref{tab:license} details the license for each corpus we used in the paper:  

\begin{table}[h]
\small
\centering
\begin{tabular}{ll}
\toprule
\textbf{Dataset} & \textbf{License}  \\
\midrule
The Pile & MIT \\
The RefinedWeb &  ODC-By 1.0 \\
ExpertQA & MIT \\
\bottomrule
\end{tabular}
\caption{Datasets' licenses}
\label{tab:license}
\end{table}
\begin{table*}[t]
\centering
\small
\begin{tabular}{ll|ccc|c}
\toprule
\textbf{Model} & \textbf{Variant} & \textbf{Neutral} & \textbf{True (T)} & \textbf{False (F)} & \textbf{Fact. Accuracy $\left(=\frac{\text{T}}{\text{T+F}}\right)$}\\
\midrule
\multirow{3.5}{50pt}{\textbf{GPT-J 6B}}
& Right & 62.4\% &	11.3\% & 26.3\% & 30.0\% \\
& Wrong& 48.8\% & ~~5.4\% & 45.8\% & 10.5\% \\
\cmidrule{2-6}
& All (Weighted) &59.5\% & 10.0\% &	30.5\% & 24.8\% \\
\midrule
\multirow{3.5}{50pt}{\textbf{OPT-66B}}
& Right & 54.1\% & 21.4\% & 24.5\% & 46.6\% \\
& Wrong & 55.1\% & ~~2.1\% & 42.8\% & ~~4.6\% \\
\cmidrule{2-6}
& All (Weighted) & 54.3\% & 17.7\% & 28.4\% & 38.8\% \\
\bottomrule
\end{tabular}

\caption{Manual factuality annotation results for OPT-66B and GPT-J 6B. For each model, we present the results per \textit{right} and \textit{wrong} subsets. Bottom row shows the weighted average between the \textit{right} and \textit{wrong} variants w.r.t to the \textit{right}/\textit{wrong} pairs of \wikimethod.
}
\label{tab:factuality_generation_app}
\end{table*}

\section{Prompts for Contradictions Generation}\label{app:prompts}
 We prompted the model to generate multiple candidate completions, For each of the five error types: entity (Table \ref{tab:prompt-entity}), circumstance (Table \ref{tab:prompt-circ}), coreference (Table \ref{tab:prompt-coref}), predicate (Table \ref{tab:prompt-pred-1} and \ref{tab:prompt-pred-2}) and link (Table \ref{tab:prompt-link}). The prompts are concatenated to a given a completion and its near context, with the exception of link-prompt where only the completion is given (we found that the instruct model tends to repeat the context when it's appended to this particular prompt). The prompts instruct the model to first plan its local edits, and then generate the contradiction. 

\begin{table*}[t!]
\small
\centering
\begin{tabular}{ll}
\toprule
\textbf{Type} & \textbf{Prompt} \\
\midrule
Entity &  Given a context and a completion, write diverse alternative completions that contradict \\& the original completion meaning. \\&
First, identify if the completion contains an entity. Then, write the contradiction by \\& modifying an entity or it's property, add additional modifications if necessary. \\& Make sure the changes you make are minimal (so only change necessary details to make \\&  the sentence plausible). Do not modify dates or quantities. \\&
\#\# \\&
Context: "Sorry" is a song by American singer Madonna from her tenth studio album \\& Confessions on a Dance Floor (2005). It was written and produced by Madonna and \\& Stuart Price, and released as the second single from the album on February 7, 2006.\\& It later appeared on Celebration, her 2009 greatest hits album. An uptempo dance song,\\& " Sorry " was one of the first tracks developed for the album and had numerous remix\\& treatments before the ultimate version of the track was finalized.\\&
Completion: One of the remixes was done by the known band the Pet Shop Boys,\\& featuring added lyrics by the band.\\&
1. Change: "Pet Shop Boys" to "Maddona". \\&
Contradiction: One of the remixes was done by the known singer Maddona,\\& featuring added lyrics by the singer.
2. Change: "Pet Shop Boys" to "Depeche Mode".\\&
Contradiction: One of the remixes was done by the known band Depeche Mode,\\& featuring added lyrics by the band. \\&
3. Change: "known" to "unfamiliar".\\&
Contradiction: One of the remixes was done by the unfamiliar band Pet Shop Boys,\\& featuring added lyrics by the band.\\&
4. Change: "Pet Shop Boys" to "the Killers".\\&
Contradiction: One of the remixes was done by the known band the Killers,\\& featuring added lyrics by the band.\\&
        \#\#\\ &
        Context: \{context\}\\&
        Completion: \{completion\}\\
\bottomrule
\end{tabular}
\caption{Prompt for entity-errors generation}
\label{tab:prompt-entity}
\end{table*}

\begin{table*}[t!]
\small
\centering
\begin{tabular}{ll}
\toprule
\textbf{Type} & \textbf{Prompt} \\
\midrule
Circumstance & Given a context and a completion, write diverse alternative completions that contradict the \\& original completion meaning. \\&
First, identify if the completion describes the circumstances of an event (location or time). If \\& circumstances are mentioned, modify it to contradict the completion. Do not add time or location if \\& they didn't appear in the original completion. Make sure the changes you make are minimal.\\&
\#\#\\&
Context: The kingdom had been in long gradual decline since the early 13th century. Had Pagan \\& possessed a stronger central government, the collapse could have been temporary, and the country \\& "could have risen again". But the dynasty could not recover, and because the Mongols refused to fill \\& the power vacuum, no viable center emerged in the immediate aftermath. As a result, several minor \\& states fought it out for  supremacy for the better part of the 14th century. \\&
Completion: It was only in the late 14th century that two relatively strong powers emerged in the \\& Irrawaddy basin, restoring some semblance of normalcy.\\&
1. Change: "14th" to "15th".\\&
Contradiction: It was only in the late 15th century that two relatively strong powers emerged in the \\&Irrawaddy basin, restoring some semblance of normalcy.
2. Change: "Irrawaddy" to "Chindwin".\\&
Contradiction: It was only in the late 14th century that two relatively strong powers emerged in the \\&Chindwin basin, restoring some semblance of normalcy.\\&
3. Change: "late" to "mid".\\&
Contradiction: It was only in the mid 14th century that two relatively strong powers emerged in the \\& Irrawaddy basin, restoring some semblance of normalcy.\\&
\#\#\\&
Context: \{context\}\\&
Completion: \{completion\}
\\
\bottomrule
\end{tabular}
\caption{Prompt for circumstance-errors generation}
\label{tab:prompt-circ}
\end{table*}

\begin{table*}[t!]
\small
\centering
\begin{tabular}{ll}
\toprule
\textbf{Type} & \textbf{Prompt} \\
\midrule
Coreference &  Given a context and a completion, write diverse alternative completions that contradict \\ & the  original completion meaning. First, decide if the completion contains a pronoun \\& (such as: he, she, it, they, his, her, its, theirs...) and write the entity it refers to. \\ & Write the contradiction by modifying the pronoun to contradict the original coreference. \\& \#\#\\ & Context: His stance in favor of prohibition cost him the votes of four legislators in his \\ &  own party and the seat went to Republican William O. Bradley. Six years later \\ & Beckham secured the seat by popular election, but he lost his re-election bid largely  \\ & because of his pro-temperance  views and his opposition to women's suffrage. \\ & Completion: Though he continued to play an active role in state politics for \\& another two decades, he never returned to elected office, failing in his gubernatorial   \\& bid in 1927 and his senatorial campaign in 1936.\\ & 1. Pronoun: he\\ & Change: "he" to "Bradley".\\ &
Contradiction: Though Bradley continued to play an active role in state politics for \\ & another two decades, he never returned to elected office, failing in his gubernatorial \\ & bid in 1927 and his senatorial campaign in 1936.\\
 & 2. Pronoun: he \\ & Change: "he" to "Bradley".\\ & Contradiction: Though he continued to play an active role in state politics for \\ & another two decades, Bradley never returned to elected office, failing in his \\ & gubernatorial  bid in 1927 and his senatorial campaign in 1936.\\ & 3. Pronoun: his\\ & Change: "his" to "Bradley's".\\ & Contradiction: Though he continued to play an active role in state politics for \\ & another two decades, he never returned to elected office, failing in Bradley's \\ & gubernatorial bid in 1927 and his senatorial campaign in 1936.\\&
 \#\# \\&
 Context: The early 6th century saw another queen ruling the city, known only as the \\& "Lady of Tikal", who was very likely a daughter of Chak Tok Ich 'aak II.\\& 
 Completion: She seems never to have ruled in her own right, rather being partnered \\&  with other rulers.\\& 
 1. Pronoun: She\\& 
 Change: "She" to "He" and "her" to "his".\\& 
 Contradiction: He seems never to have ruled in his own right, rather being partnered \\&  with other rulers.\\& 
 2. Pronoun: She\\&
 Change: "She" to "The king" and "her" to "his".\\& 
 Contradiction: The king seems never to have ruled in his own right, rather \\& being partnered with other rulers.\\&
3. Pronoun: She\\&
Change: "She" to "Chak Tok Ich".\\&
Contradiction: Chak Tok Ich seems never to have ruled in her own right, rather \\& being partnered with other rulers.\\&
        \#\#\\ &
        Context: \{context\}\\&
        Completion: \{completion\}\\
\bottomrule
\end{tabular}
\caption{Prompt for coreference-errors generation}
\label{tab:prompt-coref}
\end{table*}

\begin{table*}[t!]
\small
\centering
\begin{tabular}{ll}
\toprule
\textbf{Type} & \textbf{Prompt} \\
\midrule
Predicate &  Given a context and a completion, write diverse alternative completions, that contradict the original \\& completion meaning by modifying verbs. \\&
First, Identify a verb in the original completion, and then write the contradiction by   modifying it. Make sure \\& the contradictions are grammatically correct, fluent and consistent. Make any necessary additional \\& modifications to ensure that. \\&
\#\#\\&
Context: Homarus gammarus is a large crustacean, with a body length up to 60 centimetres (24 in) and \\& weighing up to 5 – 6 kilograms (11 – 13 lb), although the lobsters caught in lobster pots are usually \\& 23 – 38 cm (9 – 15 in) long and weigh 0.7 – 2.2 kg (1.5 – 4.9 lb).\\&
Completion: Like other crustaceans, lobsters have a hard exoskeleton which they must shed in order to grow, \\&  in a process called ecdysis (moulting).\\&
1. Change: "shed" to "retain". Additional changes: "in order to grow" to "in order to survive".
\\&
Contradiction: Like other crustaceans, lobsters have a hard exoskeleton which they must retain in order to \\&  survive, in a process called ecdysis (moulting).
\\&
2. Change: "grow" to "maintain their size". \\&
Contradiction: Like other crustaceans, lobsters have a hard exoskeleton which they must shed in order to \\& maintain their size, in a process called ecdysis (moulting).
\\&
3. Change: "shed" to "keep". Additional changes: "in order to grow" to "in order to strengthen".\\& 
Contradiction: Like other crustaceans, lobsters have a hard exoskeleton which they must keep in order to \\& strengthen, in a process called ecdysis (moulting).\\&
\#\# \\&
Context: The ridge offered a natural avenue of approach to the airfield, commanded the surrounding area \\&  and was almost undefended. Edson and Thomas tried to persuade Vandegrift to move forces to defend \\& the ridge, but Vandegrift refused, believing that the Japanese were more likely to attack along the coast. \\&
Completion: Finally, Thomas convinced Vandegrift that the ridge was a good location for Edson's Raiders \\& to rest from their actions of the preceding month. \\&
1. Change: "rest" to "keep up". \\&
Contradiction: Finally, Thomas convinced Vandegrift that the ridge was a good location for Edson's \\& Raiders to keep up with their actions of the preceding month. \\&
2. Change: "convinced Vandegrift" to "made Vandegrift doubt".\\&
Contradiction: Finally, Thomas made Vandegrift doubt that the ridge was a good location for Edson's \\& Raiders to rest from their actions of the preceding month.
3. Change: "rest" to "continue".\\&
Contradiction: Finally, Thomas convinced Vandegrift that the ridge was a good location for Edson's \\& Raiders to continue their actions of the preceding month.\\&
\#\#\\&
Context: According to a report titled Wolves in Sheep's Clothing, which documents the increase in \\& potentially violent, profane, and sexual content in children's programming, the Parents Television Council, \\& a watchdog media group, and fans believed the SpongeBob SquarePants episode" Sailor Mouth "was \\& an implicit attempt to promote and satirize use of profanity among children.\\&
Completion: The episode originally aired during the 2001 – 02 television season, ironically the season \\&  in which the PTC named SpongeBob SquarePants among the best programs on cable television, \\& but the report cited a repeat broadcast of the episode from 2005 to prove its point that it promoted use of \\& profanity among children. \\&
1. Change: "prove" to "refute". Additional changes: "best" to "most profane". \\&
Contradiction: The episode originally aired during the 2001 – 02 television season, ironically the season \\& in which the PTC named SpongeBob SquarePants among the most profane programs on cable television, \\& but the report cited a repeat broadcast of the episode from 2005 to refute its point that it promoted use of \\& profanity among children.\\&
2. Change: "originally aired" to "pulled off". \\&
Contradiction: The episode was pulled off from the 2001 – 02 television season, ironically the season \\& in which the PTC named SpongeBob SquarePants among the best programs on cable television, \\& but the report cited a repeat broadcast of the episode from 2005 to prove its point that it promoted use of \\& profanity among children.\\&
\#\#\\ &
Context: \{context\}\\&
Completion: \{completion\}
\\
\bottomrule
\end{tabular}
\caption{Prompt for predicate-errors generation (the rest of the prompt is in table \ref{tab:prompt-pred-2})}
\label{tab:prompt-pred-1}
\end{table*}

\begin{table*}[t!]
\small
\centering
\begin{tabular}{ll}
\toprule
\textbf{Type} & \textbf{Prompt} \\
\midrule
Predicate & Context: By Part II of the series, Shikamaru is capable of utilizing multiple shadow-based techniques at \\&  once and can lift his shadow from the ground in order to interact with physical objects; for instance, he can \\&  pierce enemies with the shadow tendrils or use them to throw weapons. Shikamaru approaches the exams \\&  with a sense of apathy; when he battles the Sunagakure ninja Temari, he defeats her \\&  but forfeits his match to her, due to his chakra being low.\\&
Completion: Despite this loss, he is the only ninja among his peers to be promoted to the rank of Chunin,\\& as the overseers of the exams were impressed by the insight and intelligence he demonstrated against Temari.\\&
1. Change: "promoted" to "demoted". Additional changes: "Despite" to "Due", "as" to "although".\\&
Contradiction: Due to this loss, he is the only ninja among his peers to be demoted to the rank of Chunin, \\& although the overseers of the exams were impressed by the insight and intelligence he demonstrated against \\& Temari.\\&
2. Change: "were impressed" to "underappreciated". Additional changes: "as" to "although".\\&
Contradiction: Despite this loss, he is the only ninja among his peers to be promoted to the rank of Chunin, \\& although the overseers of the exams underappreciated the insight and intelligence he demonstrated against \\& Temari.\\&
3. Change: "demonstrated" to "failed to demonstrate". Additional changes: "as" to "although", \\& "impressed" to "disappointed".\\&
Contradiction: Despite this loss, he is the only ninja among his peers to be promoted to the rank of Chunin, \\& although the overseers of the exams were disappointed by the insight and intelligence he failed to \\& demonstrate against Temari.\\&
\#\#\\ &
Context: \{context\}\\&
Completion: \{completion\}
\\
\bottomrule
\end{tabular}
\caption{Prompt for predicate-errors generation (continue of the prompt in table \ref{tab:prompt-pred-1})}
\label{tab:prompt-pred-2}
\end{table*}

\begin{table*}[t!]
\small
\centering
\begin{tabular}{ll}
\toprule
\textbf{Type} & \textbf{Prompt} \\
\midrule
Link & Given a sentence, write contradictory sentences by modifying a temporal link. \\&
First, identify a link between events, and then modify it. Make sure the contradictions are grammatically \\& correct and fluent.  
If no such link exists, answer "NA". \\&
\#\# \\&
Sentence: Prior to filming, a week was spent reinforcing the roof of the liquor store to ensure it would not \\& collapse if it were to be intruded by a group of fans.\\&
1. Change: "prior to" to "after".\\&
Contradiction: After filming, a week was spent reinforcing the roof of the liquor store to ensure it would not \\& collapse if it were to be intruded by a group of fans.\\& 
\#\#\\&
Sentence: Lewis McAllister, a businessman in Tuscaloosa, Alabama, was the first Republican to serve in the \\& Mississippi House of Representatives since Reconstruction, 1962-1968; he resided in Meridian prior to 1971.\\&
1. Change: "prior to" to "after".\\&
Contradiction: Lewis McAllister, a businessman in Tuscaloosa, Alabama, was the first Republican to serve \\& in the Mississippi House of Representatives since Reconstruction, 1962-1968; he resided in Meridian \\& after 1971.\\&
2. Change: "since" to "before"\\&
Contradiction: Lewis McAllister, a businessman in Tuscaloosa, Alabama, was the first Republican to serve \\& in the Mississippi House of Representatives before Reconstruction, 1962-1968; he resided in Meridian prior \\& to 1971.\\&
\#\#\\&
Sentence: The decline of the railroad industry caused significant job losses, resulting in a population decline \\& as workers left for other areas.\\&
1. Change: "caused" to "caused by".\\&
Contradiction: The decline of the railroad industry, caused by significant job losses, resulting a \\& population decline as workers left for other areas.\\&
2. Change: "resulting" to "was the result of".\\&
Contradiction: The decline of the railroad industry caused significant job losses, was the result of a population\\& decline, as workers left for other areas.\\&
\#\#\\&
Sentence: \{completion\}
\\
\bottomrule
\end{tabular}
\caption{Prompt for link-errors generation}
\label{tab:prompt-link}
\end{table*}

\end{document}